\documentclass[letterpaper]{article} 
\usepackage{aaai2026}  
\usepackage{times}  
\usepackage{helvet}  
\usepackage{courier}  
\usepackage[hyphens]{url}  
\usepackage{graphicx} 
\usepackage{natbib}  
\usepackage{caption} 
\usepackage{algorithm}
\usepackage{algorithmic}
\usepackage{multirow}
\usepackage{pifont}
\newcommand{\cmark}{\ding{51}} 
 
\usepackage{makecell}
\usepackage{booktabs}
\usepackage{newfloat}
\usepackage{listings}
\usepackage{arydshln}
\usepackage{amsmath}
\usepackage{bibentry}
\DeclareCaptionStyle{ruled}{labelfont=normalfont,labelsep=colon,strut=off} 
\floatstyle{ruled}
\newfloat{listing}{tb}{lst}{}
\floatname{listing}{Listing}
\title{Rethinking Surgical Smoke: A Smoke-Type-Aware Laparoscopic Video Desmoking Method and Dataset}

\author{
    Qifan Liang\textsuperscript{\rm 1,2},
    Junlin Li\textsuperscript{\rm 3},
    Zhen Han\textsuperscript{\rm 1,2}\thanks{Corresponding author.},
    Xihao Wang\textsuperscript{\rm 1,2},
    Zhongyuan Wang\textsuperscript{\rm 1,2},
    Bin Mei\textsuperscript{\rm 4}
}

\affiliations{
    \textsuperscript{\rm 1}National Engineering Research Center for Multimedia Software, School of Computer Science, Wuhan University, China\\
    \textsuperscript{\rm 2}Hubei Key Laboratory of Multimedia and Network Communication Engineering, Wuhan University, China\\
    \textsuperscript{\rm 3}School of Cyber Science and Engineering, Wuhan University, China\\
    \textsuperscript{\rm 4}Zhongnan Hospital, Wuhan University, China\\
    \{liangqifan, junlin\_li, hanzhen, wangxihao, neuromei20\}@whu.edu.cn, wzy\_hope@163.com
}

\nocopyright
\begin{document}

\maketitle

\begin{abstract}
Electrocautery or lasers will inevitably generate surgical smoke, which hinders the visual guidance of laparoscopic videos for surgical procedures. The surgical smoke can be classified into different types based on its motion patterns, leading to distinctive spatio-temporal characteristics across smoky laparoscopic videos. However, existing desmoking methods fail to account for such smoke-type-specific distinctions. Therefore, we propose the first \textbf{S}moke-\textbf{T}ype-\textbf{A}ware Laparoscopic Video Desmoking \textbf{Net}work (STANet) by introducing two smoke types: \textbf{Diffusion Smoke} and \textbf{Ambient Smoke}. Specifically, a smoke mask segmentation sub-network is designed to jointly conduct smoke mask and smoke type predictions based on the attention-weighted mask aggregation, while a smokeless video reconstruction sub-network is proposed to perform specially desmoking on smoky features guided by two types of smoke mask. To address the entanglement challenges of two smoke types, we further embed a coarse-to-fine disentanglement module into the mask segmentation sub-network, which yields more accurate disentangled masks through the smoke-type-aware cross attention between non-entangled and entangled regions. In addition, we also construct the first large-scale synthetic video desmoking dataset with smoke type annotations. Extensive experiments demonstrate that our method not only outperforms state-of-the-art approaches in quality evaluations, but also exhibits superior generalization across multiple downstream surgical tasks.
\end{abstract}

\begin{links}
    \link{Datasets}{https://simon-leong.github.io/STSVD/}
\end{links}
\section{Introduction}

Laparoscopic videos can provide real-time visual feedback from surgical procedures, serving as indispensable guidance for high-precision surgery. However, high-energy surgical tools such as electrocautery or lasers will inevitably generate surgical smoke due to the cauterization of tissue components, which severely degrades the visibility of laparoscopic video. This degradation can obscure anatomical structures and hinder clinical decision-making, highlighting the importance of laparoscopic video desmoking methods. 

The most relevant traditional technology for laparoscopic desmoking is unsupervised dehazing \cite{fan2024driving,fan2025depth} and supervised dehazing \cite{xu2023video,song2023vision,fu2025iterative,fang2025guided}, which have demonstrated generalization capabilities in various scenarios. However, due to the lack of consideration for smoke spatio-temporal characteristics, dehazing methods fail to achieve satisfactory results when dealing with varying smoky motion patterns.

\begin{figure}[t]
\centering
\includegraphics[width=0.9\columnwidth]{images/Intro.jpg} 
\caption{Visual comparison of two smoke types in surgical videos. From top to bottom, the rows represent smoky frames at successive time steps and their corresponding smoke masks.}
\label{fig1}
\end{figure}

In recent years, some researchers have focused on laparoscopic image desmoking and proposed both unsupervised methods \cite{9261326,pan2022desmoke} and supervised methods \cite{sidorov2020generative,zhang2023progressive,liu2024smoke}. More recently, a laparoscopic video desmoking method \cite{wu2024self} has been proposed, which introduces an unalignment masking strategy along with a temporal coherence regularization term to further enhance the smoke removal performance in video. However, smoke will regenerate turbulence and form a new motion pattern when it collides with the surgical cavity \cite{kim2008wavelet}. Therefore, as shown in the Fig.\ref{fig1}, surgical smoke can be classified into two
types based on its motion patterns: \textbf{Diffusion smoke} and \textbf{Ambient Smoke}. The former is local and directional, appearing
in the early stage (pre-collision) of surgical cauterizations, while the latter is global and directionless, appearing in the later stage (post-collision). These differences lead to distinctive spatio-temporal characteristics across smoky laparoscopic videos. However, existing desmoking methods have not considered the influence of smoke type, failing to achieve type-specific smoke removal.

To tackle this challenge, we introduce two smoke types into the desmoking method for the first time and propose the  \textbf{S}moke-\textbf{T}ype-\textbf{A}ware Laparoscopic Video Desmoking \textbf{Net}work (STANet) that consists of three sub-networks. Specifically, the smoky feature perception sub-network first extracts and refines smoky video features through the lightweight non-rigid trajectory attention. And then, the smoke mask segmentation sub-network jointly conducts smoke mask and smoke type predictions at the local patch level, where local predictions are subsequently aggregated into global masks of two smoke types via the attention-weighted mask aggregation mechanism. On this basis, the smokeless video reconstruction sub-network introduces adaptive deformable convolution and dilated convolution into two branches respectively for smoke removal guided by two types of smoke masks. Moreover, due to multiple cauterizations in surgical procedures, the Ambient Smoke from the later stage of previous cauterization tends to be entangled with the Diffusion Smoke from the early stage of subsequent cauterization. To address the entanglement challenges of two smoke types, we further embed a coarse-to-fine mask disentanglement module into the smoke mask segmentation sub-network, which yields more accurate disentangled masks through the smoke-type-aware cross attention between non-entangled and entangled regions. 

In addition, due to the lack of smoke type annotations in existing desmoking datasets, we also construct the first large-scale synthetic video desmoking dataset with smoke type labels. In this dataset, 120 smoky videos of 100 frames cover Diffusion Smoke, Ambient Smoke, and their entangled scenarios from 28 types of surgeries, along with the corresponding smoke-free videos and ground truth smoke masks. In summary, the contributions of this paper can be summarized as follows: 

\begin{itemize}
    \item We first classify surgical smoke into two types: Diffusion Smoke and Ambient Smoke, and propose the first smoke-type-aware laparoscopic video desmoking method to perform specially smoke removal guided by two smoke types.

    \item To address the entanglement challenges of two smoke types, we further propose a coarse-to-fine mask disentanglement module to yield more accurate disentangled masks.

    \item We construct the first large-scale synthetic video desmoking dataset with smoke type labels, including 120 paired clean-smoky videos (100 frames each) across 28 surgery types, along with the corresponding ground-truth smoke masks.

    \item Extensive experiments on synthetic and real datasets demonstrate that our method surpasses state-of-the-art dehazing and desmoking methods in quality evaluations, while exhibiting strong generalization across diverse downstream surgical tasks.

\end{itemize}

\section{Related Work}
\subsection{Unsupervised/Supervised Dehazing}

Dehazing is closely related to surgical desmoking, as both aim to restore clear visuals from degraded scenes. Existing dehazing methods based on deep learning can be broadly categorized into two groups: unsupervised and supervised. 
 
\textbf{Unsupervised dehazing methods} eliminate the need for paired clean-hazy data by leveraging non-aligned constraints to learn effective dehazing mappings. Notably, DVD \cite{fan2024driving} introduces a non-aligned regularization strategy based on flow-guided attention, and its follow-up \cite{fan2025depth} proposes a depth-centric framework jointly modeling dehazing and depth estimation, significantly enhancing results under the real-world misalignment scenario. Despite their annotation-free advantage, unsupervised dehazing methods often suffer from unstable training and performance limitations, motivating the development of supervised alternatives. 

\textbf{Supervised dehazing methods} rely on paired datasets to directly optimize dehazing model under explicit guidance, and achieve higher restoration accuracy by leveraging task-specific losses and deep network architectures. Among these, ASM (Atmospheric Scattering Model)-driven methods leverage physical scattering principles to guide supervised learning and bridge the synthetic-to-real domain gap, as exemplified by domain adaptation techniques embedding physical priors \cite{li2017aod, li2018benchmarking,guo2022image}, while ASM-free methods bypass explicit physical assumptions and instead rely on powerful model architectures to learn efficient dehazing mapping from paired data \cite{song2023vision, yuan2023lhnet,xu2023video,wu2023ridcp}. More recently, IPC-Dehaze \cite{fu2025iterative} employs iterative coding-decoding architecture for progressive refinement, and SGDN \cite{fang2025guided} uses the superior structural properties of YCbCr features to guide RGB features for real-world dehazing. However, these state-of-the-art dehazing methods often overlook the diverse spatio-temporal characteristics of surgical smoke, which limits their performance of smoke removal under the surgical scenarios.

\begin{figure*}[t]
\centering
\includegraphics[width=0.9\textwidth]{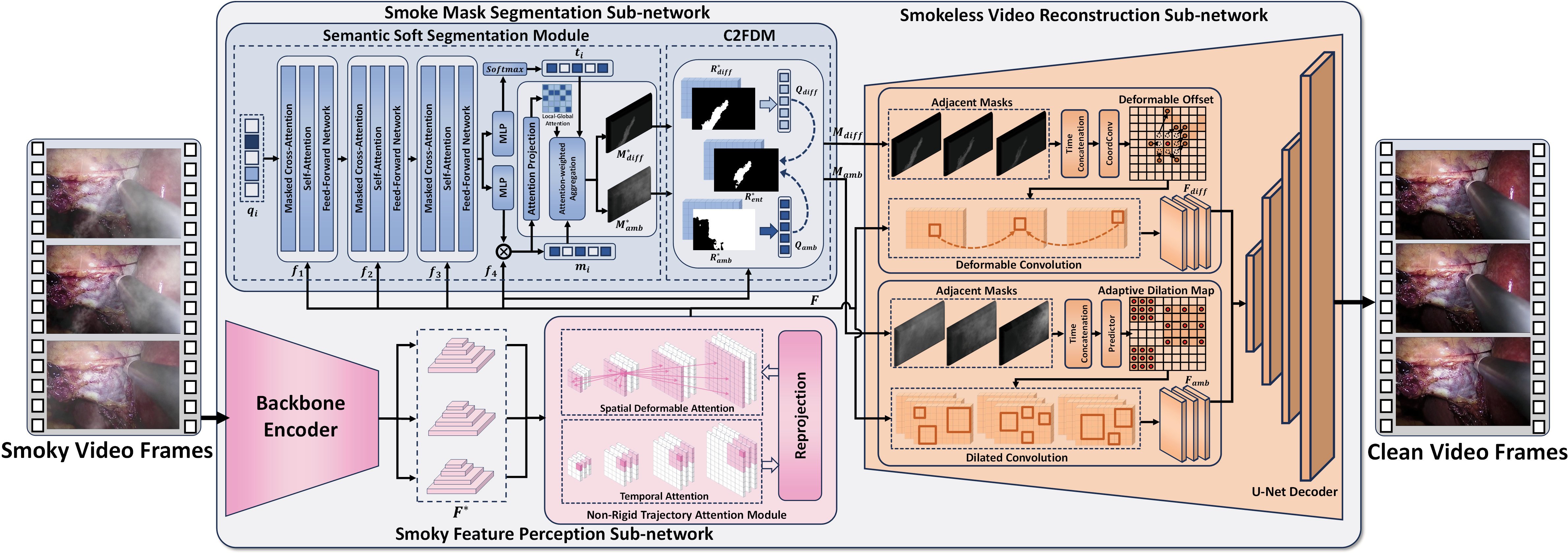} 
\caption{The overall framework of our STANet. The pink, blue and orange regions respectively denote the Smoky Feature Perception Sub-network, the Smoke Mask Segmentation Sub-network and the Smokeless Video Reconstruction Sub-network. }
\label{f2}
\end{figure*}

\subsection{Unsupervised/Supervised Desmoking}

Deep learning has driven advances in unsupervised and supervised laparoscopic desmoking. \textbf{Unsupervised desmoking methods} typically employ unpaired domain translation frameworks such as CycleGAN \cite{9261326, pan2022desmoke} to learn desmoking mappings without aligned training data. However, these methods often suffer from mode collapse and loss of details. Therefore, \textbf{Supervised desmoking methods} introduce smoke rendering engines \cite{chen2019smokegcn, holl2020phiflow,sidorov2020generative,zhang2023progressive,ma2025physliquidphysicsinformeddatasetestimating} to generate synthetic paired desmoking datasets, enabling supervised learning with more stable convergence and higher restoration fidelity. Recently, transformer-based methods like AALIDNet \cite{liu2024smoke} introduce a two-stage framework that integrates smoke mask estimation with guided embedding for smokeless feature reconstruction. SelfSVD \cite{wu2024self} introduces a video desmoking framework that leverages pre-smoke frames as pseudo ground truth to enable real-world supervision learning, and enhances desmoking performance through an unaligned masking strategy and a temporal coherence regularization term. However, existing desmoking methods still oversimplify smoke as a single type and neglect the diversity of its motion patterns, which ultimately results in unsatisfactory desmoking performance.

\section{Methodology}

As illustrated in Fig.\ref{f2}, we propose the  \textbf{S}moke-\textbf{T}ype-\textbf{A}ware Laparoscopic Video Desmoking \textbf{Net}work (STANet) for end-to-end laparoscopic video desmoking, which consists of three sub-networks. In the following sections, we provide a detailed description of each sub-network in sequence.

\subsection{Smoky Feature Perception Sub-network}

As shown in the pink region of Fig.\ref{f2}, a sequence of smoky video frames is first fed into an interchangeable ResNet-18 Backbone Encoder to extract multi-scale spatial features $F^*$. Considering the non-rigid nature of surgical smoke, we draw inspiration from the reversed temporal-spatial formulation in SODA~\cite{liu2023scotch} and introduce a lightweight non-rigid trajectory attention module to capture the deformable characteristics of $F^*$. Specifically, while preserving the core attention and reprojection mechanism of SODA, we introduce shared projection layers, a window-based attention scheme, and reductions in both the number of attention heads and the dimensionality of vectors to compress the computational load of temporal attention and spatial deformable attention.

\subsection{Smoke Mask Segmentation Sub-network}
\subsubsection{Semantic Soft Segmentation Module}

To enable soft segmentation of different types of surgical smoke, we introduce a Semantic Soft Segmentation Module (S$^3$M) within the smoke mask segmentation sub-network, as illustrated in the blue region of Fig.\ref{f2}. 

Inspired by recent work \cite{cheng2022masked}, we formulate the soft segmentation of different types of smoke as a set prediction paradigm, in which $N$ learnable queries $q_{i} (i=1\cdots N)$ are iteratively refined to act as $N$ localized smoke-type-aware experts for distinct regions of interest, and each query $q_{i}$ predicts a corresponding local smoke mask $m_{i}$ and smoke type $t_{i}$. Specifically, $q_{i}$ is first input into three cascaded segmentation blocks, each of which consists of masked cross-attention \cite{cheng2022masked}, self-attention and a feedforward network. In three blocks, $q_{i}$ sequentially interacts with the first three smaller scales of spatio-temporal features $F=\{ f_l \}_{l=1}^4$ outputted by the non-rigid trajectory attention module, and enables the iterative incorporation of multi-scale smoky features into the smoke queries representation. After passing through all three blocks, each smoke query not only interacts with the fourth larger-scale feature $f_4$ via an MLP to generate the smoke mask $m_{i}$, but is also transformed by another MLP and a softmax operation into the smoke type $t_{i}$. 

To enable the aggregation of all local masks $m_{i}$ into two smoke-type-specific global masks $M^{*}_{diff}$ and $M^{*}_{amb}$, each $m_{i}$ is further passed through a CNN-based attention projection to generate a local-global attention weight $w_{i}$. And then, under the guidance of smoke type $t_{i}$, the smoke-type-specific global smoke mask $M^{*}_{tpy}$ is computed through the attention-weighted local-global mask aggregation mechanism, as defined by the following equation:

\begin{equation}
M^{*}_{typ} = \sum_i \frac{w_i}{\sum_i w_i} \cdot m_i, \quad \{i \mid t_i = {typ}\}
\end{equation}

where the value of $typ$ is $diff$ or $amb$, representing Diffusion Smoke or Ambient Smoke respectively. 

\begin{figure}[htbp]
\centering
\includegraphics[width=\columnwidth]{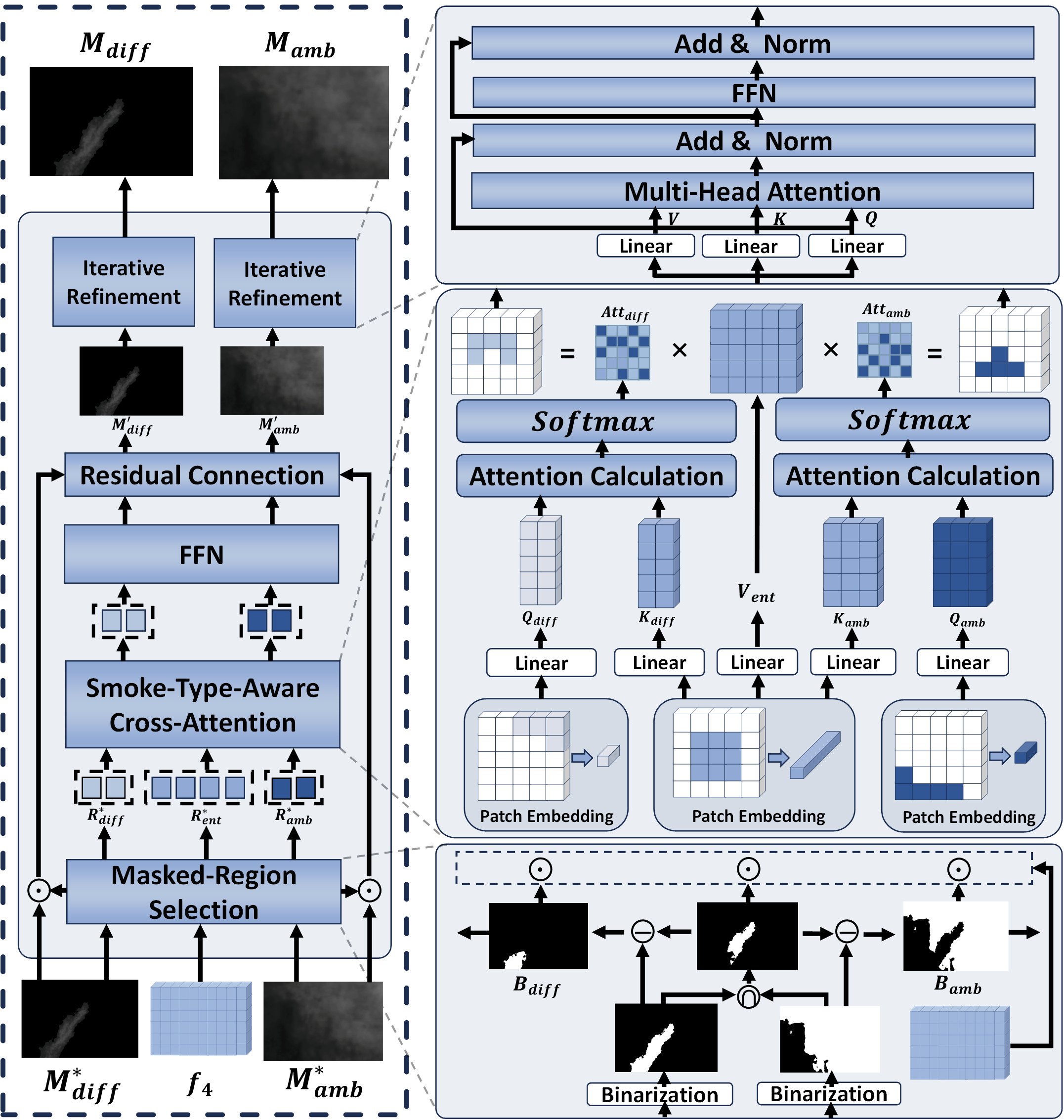}
\caption{The illustration of the Coarse-to-Fine Disentanglement Module(C2FDM).}
\label{f3}
\end{figure}

\subsubsection{Coarse-to-Fine Disentanglement Module}

Multiple cauterizations during surgery will lead to entanglement of Diffusion Smoke and Ambient Smoke generated from different cauterizations. The S$^3$M does not consider this challenge, so we further embed a Coarse-to-Fine Disentanglement Module (C2FDM) after S$^3$M to refine $M^{*}_{typ}$ and separate the entangled regions of two smoke types. 

Specifically, as shown in Fig.\ref{f3}, two coarse smoke masks without disentanglement $M^{*}_{diff}$ and $M^{*}_{amb}$ produced by S$^3$M, along with the fourth-scale feature $f_4$, are first fed into the masked-region selection sub-module of C2FDM. In this sub-module, three mutually exclusive region masks are generated from $M^{*}_{diff}$ and $M^{*}_{amb}$ via binarization and set operations, and then multiplied by $f_4$ to obtain coarse region features $R^*_{diff}$,  $R^*_{amb}$ and $R^*_{ent}$ for Diffused Smoke, Ambient Smoke and entangled regions, respectively. In the following smoke-type-aware cross attention sub-module, three region features are first passed through a patch embedding layer and a linear projection to produce smoke-type-specific queries $Q_{diff}$ and $Q_{amb}$, as well as smoke-type-specific keys $K_{diff}$ and $K_{amb}$, and a shared entangled value $V_{ent}$. On this basis, the feature of entangled region is disentangled into two smoke types through two cross attentions between non-entangled and entangled regions. Subsequently, these two smoke-type-specific features are reconstructed into two masks of entangled regions through a feedforward network $FFN$, which are fused with two masks of non-entangled regions in $M^{*}_{diff}$ and $M^{*}_{amb}$ via residual connections to produce two smoke-type-specific masks of all regions $M^{\prime}_{diff}$ and $M^{\prime}_{amb}$.The above computation process is as follows:

\begin{equation}
M^{\prime}_{typ} = FFN\left(\mathrm{Softmax}\left( \frac{Q_{typ} K_{typ}^\top}{\sqrt{d}} \right) V_{ent}\right) + M^*_{typ}\cdot B_{typ},
\end{equation}

where $B_{typ}$ represents the smoke-type-specific binary mask corresponding to non-entangled region, which is generated from the masked-region selection sub-module and used to extract the non-entangled region in $M^{*}_{typ}$.

Subsequently, the coarse masks $M^{\prime}_{diff}$ and $M^{\prime}_{amb}$ are further passed into an iterative refinement block for optimization. In detail, the refinement block consists of a multi-head attention layer with four self-attention heads, along with normalization layers and feedforward networks, which aim to progressively refine the smoke features and ultimately produce fine masks $M_{diff}$ and $M_{amb}$.

\subsection{Smokeless Video Reconstruction Sub-network}

Considering the locality and directionality of Diffusion Smoke, as well as the globality and non-directionality of Ambient Smoke, we employ a dual-branch desmoking process guided by smoke-type-specific masks within the smokeless video reconstruction sub-network.

For Diffusion Smoke, as illustrated in the upper branch of the orange region in Fig.\ref{f2}, we adopt deformable convolution to perform feature desmoking along the smoke diffusion path. To leverage the temporal coherence of smoke, we first concatenate Diffusion Smoke masks from adjacent frames to form a temporal composite mask $\bar{M}_{diff}$. $\bar{M}_{diff}$ is subsequently processed by a lightweight CoordConv layer \cite{liu2018intriguing} (with 2 added coordinate channels) and a channel-reduction attention $\mathcal{A}_{attn}$ to produce an 18-channel offset field that adaptively emphasizes informative spatial displacements:

\begin{equation}
\Delta_{offset} = \mathcal{A}_{attn}\left(CoordConv\bigl(\bar{M}_{diff})\right),
\end{equation}

Subsequently, a 3×3 deformable convolution layer is applied using the offset $\Delta_{offset}$ to extract aligned features along the smoke diffusion trajectories.

\begin{equation}
F_{diff} = DeformConv\left(F, \Delta_{offset}\right).
\end{equation}

For Ambient Smoke, as illustrated in the lower branch of the orange region in Fig.\ref{f2}, we adopt an adaptive dilated convolution to accommodate the global and non-uniform smoke distribution. Specifically, we first concatenate Ambient Smoke masks from adjacent frames to obtain a temporal composite mask $\bar{M}_{amb}$, which is then fed into a CNN-based predictor to estimate $K$ adaptive dilation sampling position maps $map_k$ corresponding to $K$  dilation rates $rate_k$. Then, we apply parallel 3×3 dilated convolutions $DilatConv$ with all dilation rates and fuse their outputs:

\begin{equation}
F_{amb} = \sum_{k=1}^{K}  DilatConv\left(F, rate_k, map_k\right).
\end{equation}

where $K=3$ and  $rate_1, rate_2,rate_3$ are set to 1, 2, 3 respectively. Finally, the features $F_{diff}$ and $F_{amb}$ from two branches are routed to a U-Net decoder, which reconstructs the smokeless video frame by fusing multi-scale features through progressive upsampling. In addition, it is worth noting that when only one type of smoke appears in the video, the corresponding branch will be activated exclusively to save unnecessary computational cost.

\subsection{Overall Training Loss}

Following recent works \cite{cheng2022masked,liu2024smoke}, our STANet is optimized using a multi-task loss $\mathcal{L}_{mul}$ to jointly supervise smoke mask segmentation, smoke type classification, and smokeless video reconstruction tasks. 

To tackle the challenge of smoke mask details prediction, we introduce an additional Smoke High-frequency Wing Loss (SHWL) $\mathcal{L}_{shwl}$ to emphasize high-frequency information of masks via adaptive gradient modulation. Specifically, we first extract the high-frequency of normalized ground truth mask $M_{GT}$ and predicted mask $M_{typ}$ using a $3 \times 3$ high-pass filter, and then compute the absolute high-frequency error $\epsilon$ between them. Finally, we refer to the wing loss \cite{bedard2025towards} to optimize $\epsilon$, which can better focus on the overlooked small error of smoke mask. 
 
 To further adapt to varying smoke density, an exponential modulation factor is introduced to adjust $\mathcal{L}_{shwl}$:

\begin{equation}
\phi = 1 + \lambda_g (e^{M_{GT}} - 1)
\end{equation}

where $\lambda_g=2.0$ is empirically set to balance gradient sensitivity, and $\phi$ increases the penalty in dense smoke regions ($M_{GT} \rightarrow 1$) while reducing overfitting in sparse regions ($M_{GT} \rightarrow 0$). The total loss is calculated as follows:
\begin{equation}
\mathcal{L}_{total} = \mathcal{L}_{mul} + \phi\mathcal{L}_{shwl}  .
\end{equation}

\section{Smoke-Type-Specific Desmoking Dataset}

Due to the lack of smoke type annotations in existing desmoking datasets, we construct a large-scale \textbf{S}moke-\textbf{T}ype-\textbf{S}pecific \textbf{V}ideo \textbf{D}esmoking (STSVD) dataset, including 120 videos (100 frames each, 720$\times$1080 resolution) with semantic labels for Diffusion Smoke, Ambient Smoke, and their entanglement across 28 types of surgical scenarios.

\begin{figure}[htbp]
    \centering
    \includegraphics[width=\columnwidth]{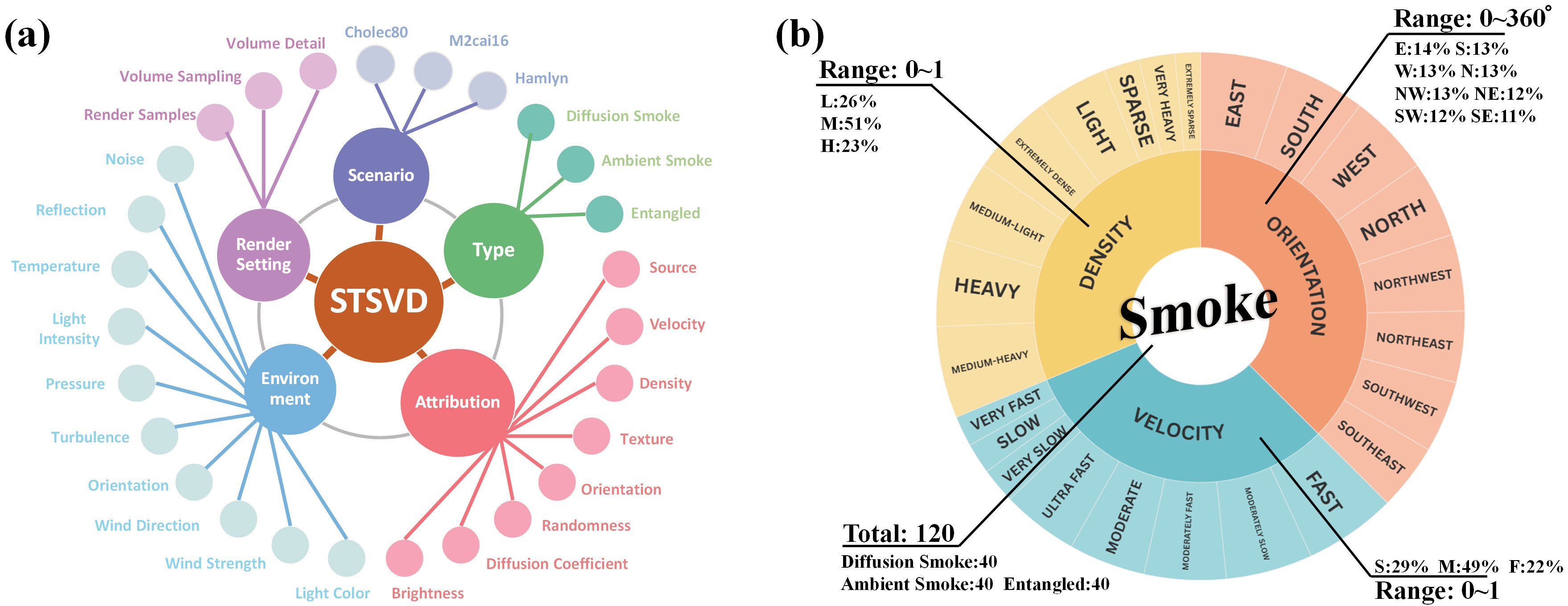}
    \caption{Statistics of the proposed STSVD dataset.}
    \label{f4}
\end{figure}

As illustrated in Fig.\ref{f4}(a), in order to ensure a realistic smoke simulation, STSVD considers five dimensions: surgical scenario, smoke type, smoke attribute, environment parameter and render setting, which are divided into 27 sub-dimensions. In addition, Fig.\ref{f4}(b) further illustrates the subdivisions of three major smoke attributes: density, orientation and velocity, which follow an empirically established distribution of real-world surgical scenarios \cite{hong2023mars}. More details of STSVD are provided in the code.

\begin{table*}[t]
  \renewcommand{\arraystretch}{1.28}
  \setlength{\tabcolsep}{3pt}
  \centering
  \resizebox{\textwidth}{!}{%
    \begin{tabular}{c|c|c|ccc|>{\centering\arraybackslash}p{1.5cm}
    >{\centering\arraybackslash}p{1.5cm}
    >{\centering\arraybackslash}p{1.5cm}|cccc|ccc}
    \Xhline{2pt}
    \multirow{2}{*}{\textbf{Settings}} & \multirow{2}{*}{\textbf{Methods}} & \multirow{2}{*}{\textbf{Venue}} 
    & \multicolumn{3}{c|}{\textbf{STSVD}(synthetic dataset)} 
    & \multicolumn{3}{c|}{\textbf{Vivo}(paired real-world dataset)}
    & \multicolumn{4}{c|}{\textbf{STSVD-R}(unpaired real-world dataset)}
    & \multicolumn{3}{c}{\textbf{Complexity}} \\
    \cline{4-16}
    & & & PSNR$\uparrow$ & SSIM$\uparrow$ & LPIPS$\downarrow$
    & PSNR$\uparrow$ & SSIM$\uparrow$ & LPIPS$\downarrow$
    & TOPIQ$\uparrow$ & Q-align$\uparrow$ & MANIQA$\uparrow$ & MUSIQ$\uparrow$
    & Params & FLOPS & Times \\
    \Xhline{1pt}

    \multirow{2}{*}{\makecell{Unsupervised\\Dehazing}}
    & DVD \cite{fan2024driving} & CVPR'24 & 26.1833 & 0.8945 & \underline{0.0436} & 22.3465 & 0.7820 & 0.3028 & 0.2805 & 3.0083 & 0.1512 & 36.7322 & 15.37M & 101.30G & 0.73s \\
    & DCL \cite{fan2025depth}& AAAI'25 & 27.8133 & 0.9362 & 0.1132 & 21.6424 & 0.8341 & \underline{0.1493} & 0.2985 & 3.0115 & 0.1813 & 38.7317 & 11.38M & 73.40G & 0.47s \\
    \hline

    \multirow{7}{*}{\makecell{Supervised\\Dehazing}}
    & AODNet \cite{li2017aod}& ICCV'17 & 26.6464 & 0.9520 & 0.0775 & 22.1000 & 0.8480 & 0.1820 & 0.3058 & 2.9501 & 0.1700 & 37.1496 & 0.02M & 1.36G & 0.32s \\
    & MapNet \cite{xu2023video}& CVPR'23  & 26.7672 & 0.9156 & 0.1014 & 22.5223 & 0.8607 & 0.1653 & 0.3068 & 3.0669 & 0.1068 & 40.0033 & 28.75M & 261.20G & 0.49s \\
    & RIDCP \cite{wu2023ridcp}& CVPR'23  & 29.0814 & 0.9581 & 0.0629 & 21.4168 & 0.8363 & 0.2100 & 0.3025 & 3.0106 & 0.1812 & 40.1816 & 28.72M & 182.69G & 0.72s \\    
    & DehazeFormer \cite{song2023vision}& TIP'23  & 32.5400 & 0.9517 & 0.0463 & \underline{23.2910} & \underline{0.8648} & 0.1569 & 0.3082 & 3.1042 & 0.1874 & 39.5138 & 28.98M & 11.16G & 0.07s \\
    & IPCDehaze \cite{fu2025iterative}& CVPR'25  & 29.8004 & 0.9629 & 0.0506 & 22.1459 & 0.8504 & 0.1823 & 0.3037 & 3.0472 & 0.1725 & 38.8480 & 43.16M & 665.42G & 4.16s \\
    & SGDN \cite{fang2025guided}& AAAI'25 &  \underline{32.6625} & \underline{0.9674} & 0.0483 & 23.1528 & 0.8631 & 0.1662 & 0.3056 & 3.0725 & 0.1855 & \underline{40.1860} & 11.09M & 53.40G & 0.84s \\
    \hline

    \multirow{2}{*}{\makecell{Unsupervised\\Desmoking}}
    & DCP-P2P \cite{9261326}& IEEE Access'20  & 23.1149 & 0.8937 & 0.2218 & 21.4891 & 0.8409 & 0.2399 & 0.2202 & 2.5458 & 0.1847 & 30.9309 & 54.42M & 194.00G & 0.82s \\
    & Desmokelap \cite{pan2022desmoke}& IJCARS'22  & 26.4311 & 0.9037 & 0.2217 & 23.2277 & 0.8470 & 0.2431 & 0.2273 & 2.8237 & 0.1776 & 32.5820 & 11.38M & 1571.00G & 0.65s \\
    \hline

    \multirow{5}{*}{\makecell{Supervised\\Desmoking}}
    & SSIM-PFAN \cite{sidorov2020generative}& PMLR'20  & 27.4094 & 0.9579 & 0.0841 & 22.0537 & 0.8031 & 0.1961 & 0.2844 & 2.9750 & 0.1773 & 38.2428 & 51.65M & 180.15G & 0.35s \\
    & PFAN \cite{zhang2023progressive}& PRCV'23 &  26.0183 & 0.8827 & 0.2031 & 21.0142 & 0.8218 & 0.2391 & 0.2458 & 2.7544 & 0.1692 & 32.3478 & 54.41M & 71.80G & 0.07s \\
    & AALIDNet \cite{liu2024smoke}& ROBIO'24  & 26.5064 & 0.9226 & 0.2125 & 21.9999 & 0.8171 & 0.3410 & 0.2780 & 2.4094 & \underline{0.1895} & 34.0062 & 33.11M & 7.88G & 0.03s \\
    & SelfSVD \cite{wu2024self}& ECCV'24  & 29.5132 & 0.9588 & 0.0842 & 22.1343 & 0.8376 & 0.1980 & \underline{0.3154} & \underline{3.1195} & 0.1784 & 38.4157 & 27.70M & 996.00G & 0.18s \\
    & \textbf{Ours} & --  & \textbf{33.5345} & \textbf{0.9733} & \textbf{0.0402} & \textbf{23.8427} & \textbf{0.8813} & \textbf{0.1446} & \textbf{0.3271} & \textbf{3.1328} & \textbf{0.1914} & \textbf{40.2063} & 25.15M & 174.13G & 0.21s \\
    \Xhline{2pt}
    \end{tabular}}
    \caption{The qualitative comparison between different methods. $\downarrow$ indicates that lower values are better, while $\uparrow$ indicates that higher values are better. The best results are highlighted in bold, and the second-best results are underlined.}
    \label{com}
\end{table*}


\begin{figure*}[t]
  \centering
  \includegraphics[width=\textwidth]{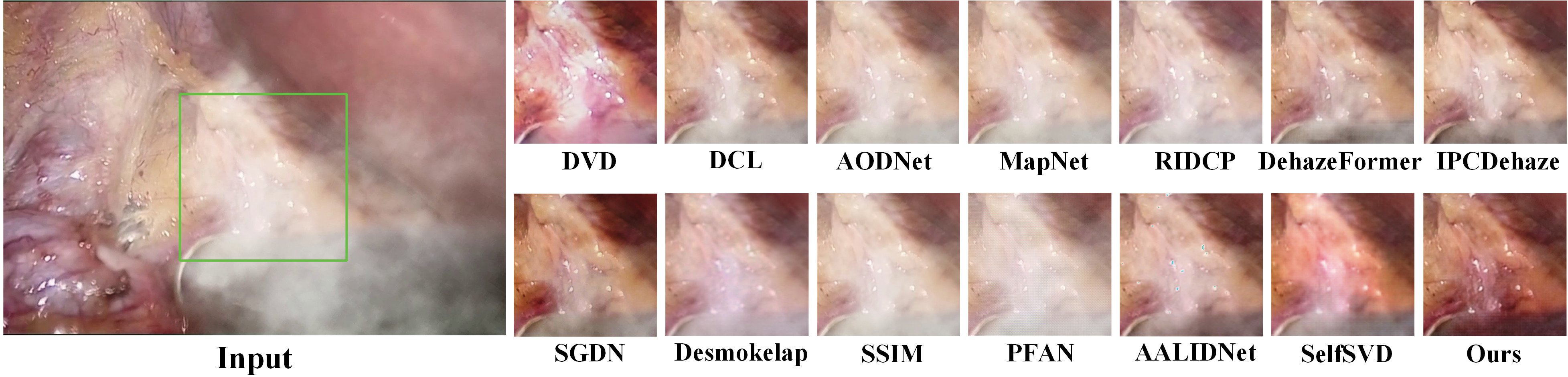}
  \caption{The qualitative comparison between different methods. More results are given on the paper website.}
  \label{fig_com}
\end{figure*}

\section{Experiments}

\subsection{Datasets and Metrics}

\subsubsection{Training Datasets} 
In comparison experiments, we use STSVD as the training dataset. In ablation experiments, we use PSv2rs \cite{liu2024smoke} as an additional training dataset to analyze the impact of different training datasets, thereby demonstrating the superiority of STSVD. PSv2rs is the largest existing open-source synthetic desmoking dataset, which consists of 54420 smoky frames with 256$\times$256 resolution. It should be noted that due to the lack of smoke type labels, PSv2rs can only be used to train the baseline model without awareness of smoke type.

\subsubsection{Testing Datasets} 
To evaluate the generalization of different desmoking methods in real-world scenarios, our testing dataset includes two real-world desmoking datasets Vivo and STSVD-R in addition to synthetic dataset STSVD. Vivo \cite{xia2024new} is the only existing real-world paired desmoking dataset, which consists of 961 pairs of smoky frames extracted from 63 laparoscopic prostatectomies and is aligned using optical flow supervision, allowing for paired quantitative assessment. Considering that Vivo does not cover diverse smoke types, we have specifically constructed an new unpaired real-world dataset STSVD-R by resampling smoky videos from three surgical datasets: Cholec80 \cite{twinanda2016endonet}, M2CAI16 \cite{stauder2016tum}, and Hamlyn \cite{giannarou2014hamlyn}, which contains 24 videos (100 frames each, 720$\times$1080) categorized into diffusion, ambient and entangled smoke scenarios. 

\subsubsection{Metrics} 
For synthetic dataset with ground truth, we use PSNR, SSIM, and LPIPS~\cite{zhang2018unreasonable} as the video reconstruction quality metrics. For real-world datasets without ground truth, we use four no-reference metrics: TOPIQ~\cite{chen2024topiq}, Q-Align~\cite{wu2023q}, MANIQA~\cite{yang2022maniqa}, and MUSIQ~\cite{ke2021musiq}, which respectively evaluate semantic consistency, cross-modal alignment, local statistics, multi-scale quality, and provide a comprehensive perceptual assessment.

\subsection{Implementation Details}

Each input frame is resized to 720$\times$1080 and augmented with random cropping and photometric distortion after normalization. The number of queries $N$  in the smoke mask segmentation sub-network is set to 100. Adam with an initial learning rate of $1\times 10^{-4}$, weight decay 0.05, gradient clipping 0.01, and a polynomial decay schedule over 90K iterations are adopted. The batch size is set to 4, and PyTorch is used to implement our model with RTX 3090 GPUs.

\subsection{Comparison with State-of-the-arts}

\subsubsection{Quantitative Comparisons} Tab.\ref{com} demonstrates the objective metrics of 8 dehazing methods, 6 desmoking methods, and our method on three datasets. Compared to the best-performing dehazing methods SGDN \cite{fang2025guided} and DehazeFormer \cite{song2023vision}, our method respectively improves reference metrics PSNR, SSIM, LPIPS and no-reference metrics TOPIQ, Q-align, MANIQA, MUSIQ by an average of 0.7770 dB, 0.0156, 0.0120 and 0.0202, 0.0445, 0.0050, 0.3564, due to the precise perception and representation for spatio-temporal characteristics of surgical smoke. Compared to the best-performing desmoking method SelfSVD \cite{wu2024self}, our method also exhibits superior performance across all metrics due to the advantage of smoke-type-specific modeling under diverse surgical scenarios. Specifically, it improves reference metrics by an average of 2.8649 dB, 0.0291, 0.0487, and no-reference metrics by 0.0117, 0.0133, 0.0130, 1.7906, respectively.

\subsubsection{Qualitative Comparisons} Fig.\ref{fig_com} illustrates the qualitative
results of different methods. It can be observed that the best-performing dehazing methods SGDN and DehazeFormer suffer from large areas of residual haze and lack color details of organ tissue, because they are difficult to utilize motion pattern information of surgical smoke and temproal complementarity between smoky frames. Meanwhile, compared to our method, the best-performing desmoking method SelfSVD tends to struggle with recovering clear textures in dense and entangled smoke regions due to the lack of smoke-type consideration, which limits its adaptability to complex smoky scenarios. Overall, our method can significantly remove varying smoke and reconstruct faithful details, which demonstrates the effectiveness and generalization of the smoke-type-aware desmoking framework.

\subsubsection{Complexity Comparison}
As shown in Tab.\ref{com}, our method has certain advantages in some dimensions of complexity compared to methods with the best desmoking performance. Specifically, our method outperforms SGDN in terms of time cost, DehazeFormer in terms of parameters, and SelfSVD in terms of FLOPS. It is worth noting that our method enables a dynamic activation strategy based on smoke-type guidance to avoid redundant computations as much as possible, when only one type of smoke appears in the video.

\begin{table*}[htbp]
\renewcommand{\arraystretch}{1.2}
\setlength{\tabcolsep}{4pt}
\centering
\resizebox{\linewidth}{!}{%
\begin{tabular}{c|ccccc|ccc|ccc|cccc}
\Xhline{2pt}
\multirow{2}{*}{Index} 
& \multicolumn{5}{c|}{\textbf{Ablation Component}} 
& \multicolumn{3}{c|}{\textbf{STSVD}(synthetic dataset)} 
& \multicolumn{3}{c|}{\textbf{Vivo}(paired real-world dataset)}
& \multicolumn{4}{c}{\textbf{STSVD-R}(unpaired real-world dataset)} \\
\cline{2-16}
& \shortstack{Train} 
& S$^3$M & C2FDM & SHWL & SVRS 
& PSNR$\uparrow$ & SSIM$\uparrow$ & LPIPS$\downarrow$ 
& PSNR$\uparrow$ & SSIM$\uparrow$ & LPIPS$\downarrow$ 
& TOPIQ$\uparrow$ & Q-align$\uparrow$ & MANIQA$\uparrow$ & MUSIQ$\uparrow$ \\
\Xhline{1pt}
M1 & PSv2rs  &  &  &  &  
& 29.2978 & 0.9104 & 0.0835 
& 22.9287 & 0.8515 & 0.1971 
& 0.2449 & 2.9621 & 0.1806 & 34.6328 \\
M2      & STSVD  &  &  &  &  
& 31.1898 & 0.9577 & 0.0589 
& 23.0744 & 0.8634 & 0.1639 
& 0.3099 & 3.0540 & 0.1822 & 38.2087 \\
M3        & STSVD  & \cmark  &  &  & 
& 31.8980 & 0.9632 & 0.0545 
& 23.1243 & 0.8755 & 0.1520 
& 0.3114 & 3.0879 & 0.1849 & 39.3290 \\
M4       & STSVD  & \cmark & \cmark &  & 
& 32.7462 & 0.9683 & 0.0491 
& 23.3514 & 0.8784 & 0.1498 
& 0.3201 & 3.1207 & 0.1886 & 39. 5179\\
M5        & STSVD  & \cmark & \cmark & \cmark &  
& 33.0871 & 0.9716 & 0.0445 
& 23.5024 & 0.8789 & 0.1492
& 0.3244 & 3.1266 & 0.1891 & 39.8560 \\

\textbf{M6}    & STSVD  & \cmark & \cmark & \cmark & \cmark 
& \textbf{33.5345} & \textbf{0.9733} & \textbf{0.0402} 
& \textbf{23.8427} & \textbf{0.8813} & \textbf{0.1446} 
& \textbf{0.3271} & \textbf{3.1328} & \textbf{0.1914} & \textbf{40.2063} \\
\Xhline{2pt}
\end{tabular}%
}
\caption{The quantitative comparisons of ablation study. The checkmark (\cmark) indicates whether an ablation component is included.}
\label{t_a}
\end{table*}

\begin{figure*}[t]
  \centering
  \includegraphics[width=\textwidth]{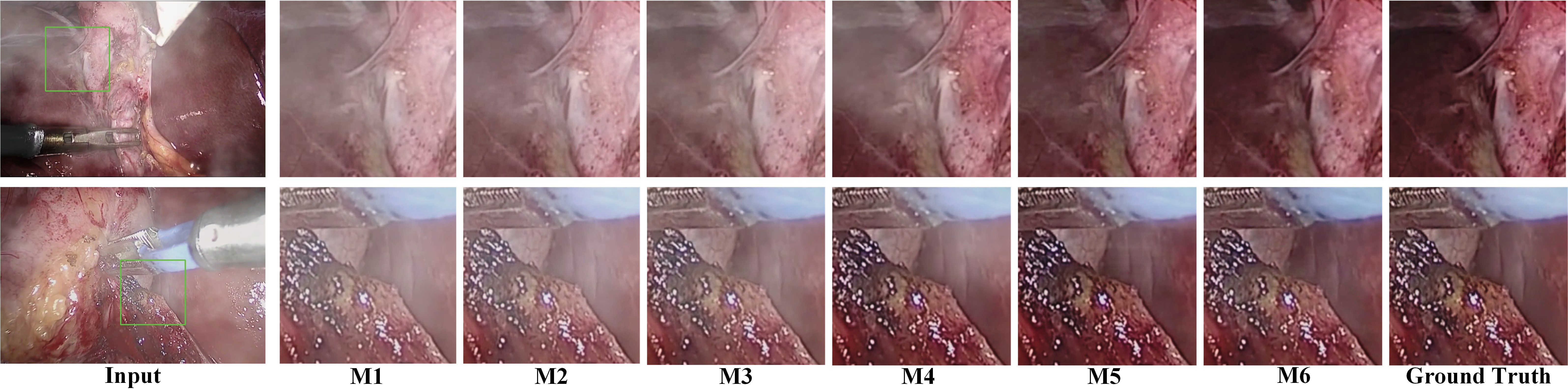}
  \caption{The qualitative comparisons of ablation study. More results are given on the paper website. }
  \label{fig_abl}
\end{figure*}

\subsection{Ablation Study}
Tab.\ref{t_a} shows the objective results of 6 ablation states, which cover various ablation components such as training dataset, network module and loss function. It can be observed that compared to M1, M2 achieves significant improvements across all reference and no-reference metrics. This is attributed to the proposed STSVD dataset, which surpasses existing PSv2rs dataset in terms of smoke fidelity and scene diversity. Based on M2, M3 further improves reference metrics PSNR, SSIM, LPIPS and no-reference metrics TOPIQ, Q-align, MANIQA, MUSIQ by an average of 0.3791 dB, 0.0088, 0.0082 and 0.0015, 0.0339, 0.0027, 1.1203, respectively. This demonstrates that the S$^3$M (semantic soft segmentation module) can jointly predict accurate smoke masks and types to provide valuable guidance for desmoking. Compared with M3, M5 improves reference metrics by an average of 0.7836 dB, 0.0059, 0.0064, and no-reference metrics by 0.0130, 0.0387, 0.0042, 0.5270, respectively. This is because the C2FDM  (coarse-to-fine disentanglement module), along with the $\mathcal{L}_{shwl}$ loss function, addresses the entanglement challenges of two smoke types and yields more accurate disentangled masks. After incorporating the SVRS (smokeless video reconstruction sub-network), M6 achieves further improvements across all metrics by leveraging two types of smoke masks to guide special desmoking on video features.

Corresponding to objective results, subjective results in Fig.\ref{fig_abl} also reflect the improvement effect of different ablation components in our method. From M1 to M6, the visual quality of smoke removal shows progressive refinement, as each component incrementally contributes to more accurate mask prediction of different smoke types and more faithful color details restoration of organ tissue.

\begin{table}[htbp]
\centering
\renewcommand{\arraystretch}{1.2}
\setlength{\tabcolsep}{3pt}
\resizebox{\linewidth}{!}{%
\begin{tabular}{c|cc|cc|cc|cc}
\Xhline{2pt}
\multirow{3}{*}{Method}
& \multicolumn{6}{c|}{\textbf{Detection}} 
& \multicolumn{2}{c}{\textbf{Segmentation}} \\
\cline{2-9}
& \multicolumn{2}{c|}{CVC-ClinicDB}
& \multicolumn{2}{c|}{CVC-ColonDB}
& \multicolumn{2}{c|}{Kvasir-SEG}
& \multicolumn{2}{c}{EndoVis18} \\
\cline{2-9}
& DSC$\uparrow$ & IoU$\uparrow$
& DSC$\uparrow$ & IoU$\uparrow$
& DSC$\uparrow$ & IoU$\uparrow$
& IoU$\uparrow$ & mcIoU$\uparrow$ \\
\Xhline{1pt}
Smoky Input
& 0.8133 & 0.7813
& 0.6597 & 0.5846
& 0.8494 & 0.8054
& 58.512 & 38.588 \\
DehazeFormer
& 0.8881 & 0.8203
& 0.6694 & 0.5916
& 0.8830 & 0.8194
& 72.768 & 49.336 \\
SGDN
& 0.8933 & 0.8315
& 0.6828 & 0.6034
& 0.8822 & 0.8190
& 74.227 & 51.795 \\
SelfSVD
& 0.8878 & 0.8254
& 0.6932 & 0.6152
& 0.8843 & 0.8205
& 68.403 & 45.463 \\
\textbf{Ours}
& \textbf{0.9024} & \textbf{0.8394}
& \textbf{0.7138} & \textbf{0.6331}
& \textbf{0.8857} & \textbf{0.8240}
& \textbf{74.814} & \textbf{51.899} \\
\Xhline{2pt}
\end{tabular}
}
\caption{Quantitative results on downstream tasks (polyp detection and instrument segmentation).}
\label{tab_ds}
\end{table}

\begin{figure}[htbp]
\centering
\includegraphics[width=\linewidth]{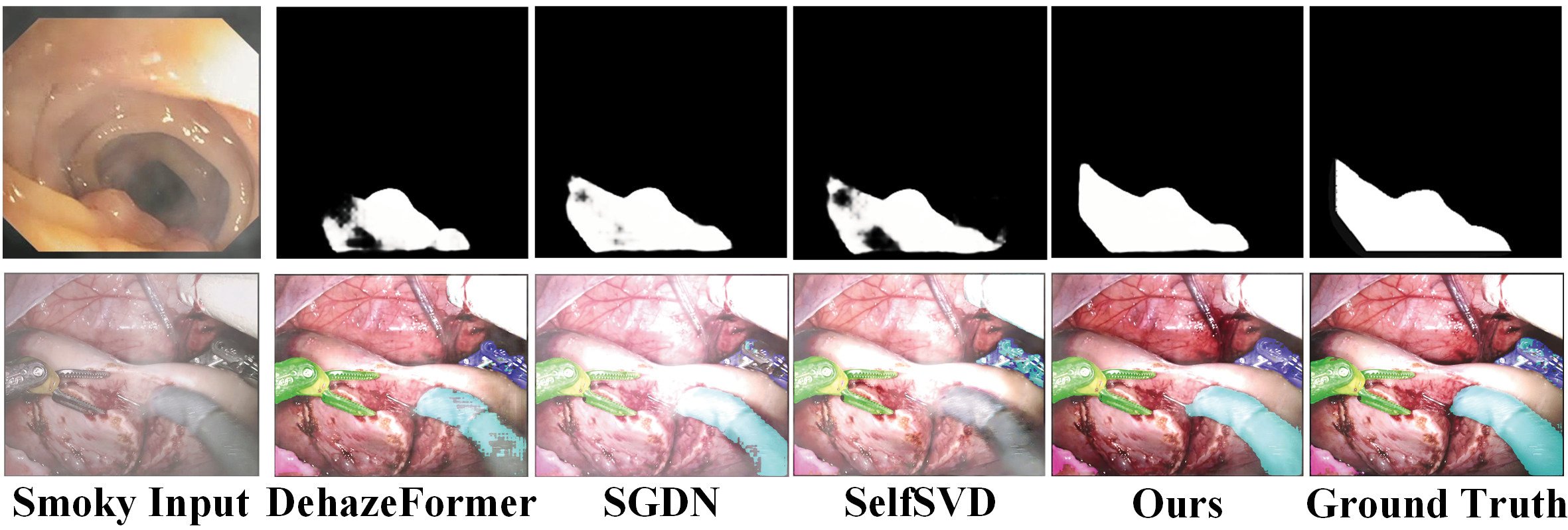}
\caption{Qualitative results on downstream tasks (polyp detection and instrument segmentation).}
\label{fig_ds}
\end{figure}

\subsection{Evaluation on Downstream Tasks}

To investigate the broader applicability of our method, we further evaluate its desmoking performance on two downstream tasks: polyp detection \cite{wei2025uninet} and surgical instrument segmentation \cite{yue2024surgicalsam}, which include three detection task testing datasets CVC-ClinicDB, CVC-ColonDB, Kvasir-SEG and one segmentation task testing dataset EndoVis18. Specifically, we respectively apply our method and three best existing methods in comparative experiments to generate desmoked inputs for downstream evaluation. As shown in Table \ref{tab_ds}, our method achieves the best performance on all datasets, surpassing the second-best method by 0.0122 in DSC and 0.0118 in IoU on the detection task, while by 0.587 in IoU and 0.104 in mcIoU on the segmentation task. Qualitative comparisons in Fig.\ref{fig_ds} further reveal that our method produces more accurate detection targets and clearer segmentation boundaries. This demonstrates that our method can effectively facilitate downstream medical tasks in smoky conditions.

\section{Conclusion}

In this paper, we propose the first smoke-type-aware laparoscopic video desmoking framework that explicitly differentiates Diffusion Smoke and Ambient Smoke. In addition, a coarse-to-fine mask disentanglement module is designed to improve mask accuracy for two types of smoke in entangled regions. To facilitate future research, we release a large-scale synthetic video desmoking dataset with smoke type annotations. Extensive experiments on synthetic and real datasets demonstrate superior performance of our method, benefiting downstream surgical tasks. 

\bigskip
\section*{Acknowledgments}
This work was supported in part by the Hubei Provincial Science and Technology Plan Project (No.~2025CSA057), and the Transformation Fund Project of Scientific and Technological Achievements of Zhongnan Hospital of Wuhan University (No.~2023CGZH-ZD006).

\bigskip

\bibliography{main}
\clearpage
\appendix
\section*{Supplementary Materials}
\addcontentsline{toc}{section}{Supplementary Materials}

In this supplementary material, Section A details the dataset construction process and includes visualization examples of the sampling. Section B and Section C presents additional qualitative results from ablation studies and comparative experiments, including STSVD, Vivo, and STSVD-R, respectively.

\section{A.Smoke-Type-Specific Desmoking Dataset}
\begin{figure}[htbp]
    \centering
    \includegraphics[width=\columnwidth]{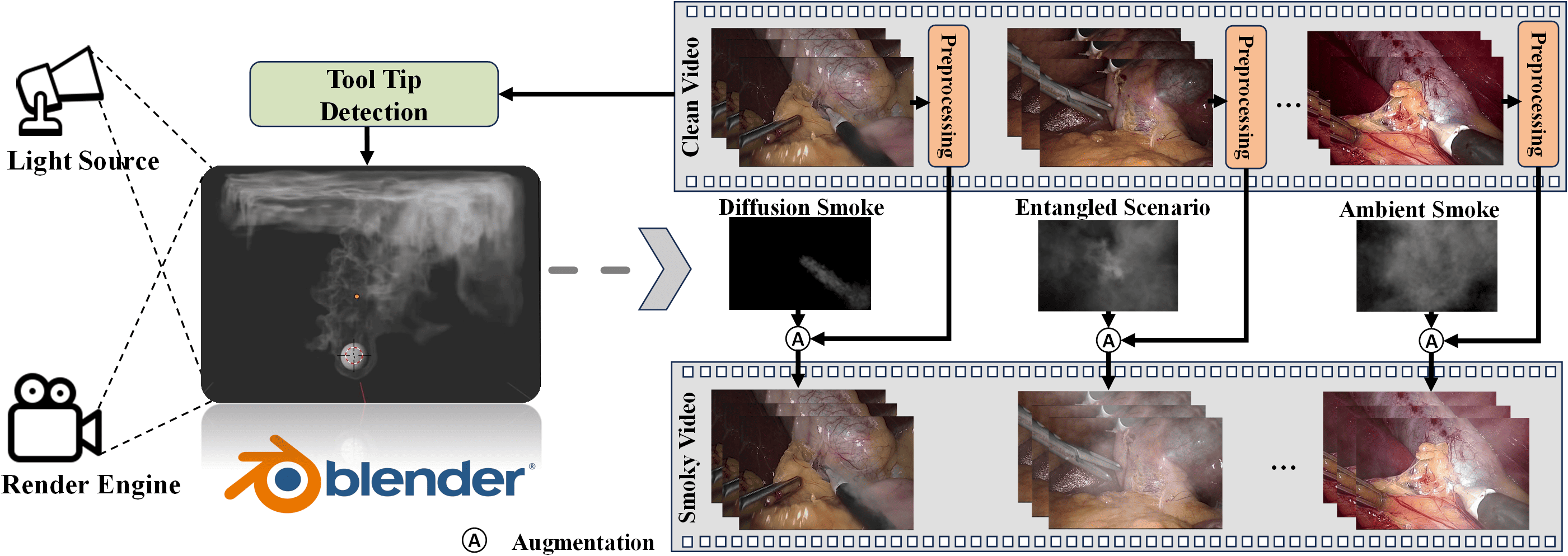}
    \caption{The construction pipeline of the proposed STSVD dataset.}
    \label{pineline}
\end{figure}

\textbf{Dataset Construction} \quad STSVD is built from high-resolution (720$\times$1080) smoke-free videos sourced from Cholec80 \cite{twinanda2016endonet}, M2CAI16 \cite{stauder2016tum}, and Hamlyn \cite{giannarou2014hamlyn}. As illustrated in Fig.\ref{pineline}, we synthesize realistic smoke using a physics-based engine following AALIDNet \cite{liu2024smoke}, simulating smoke with randomized orientation, velocity, and intensity under pressure and turbulence-driven airflow. Because the source of Diffusion Smoke should be aligned with the surgical tool tip in terms of coordinates, we need to detect the tool tip as the position of the smoke source. Specifically, we employ a CNN-based surgical tool tip detection network, comprising three convolutional layers with ReLU activation and max pooling, followed by two fully connected layers, to localize the tip of surgical tools in clean videos, which is used to define the source for smoke mask rendering. Inspired by \cite{chen2023snow}, we use pixel-wise soft compositing with adaptive transparency and optional degradations to blend synthetic smoke seamlessly. The final smoky frame at time $t$ is formulated as:

\begin{equation}
I_s(t) = I_{c}(t) + \omega \cdot \text{Aug}\left( \left( L_a - I_{c}(t) \right) \cdot \frac{M(t)}{255} \right)
\end{equation}

where $I_{c}(t)$ is the clean frame, $M(t)$ is the rendered smoke mask, $L_a = [255, 255, 255]$ denotes the atmospheric light, the blending coefficient $\omega$ is derived from the estimated smoke density in $M(t)$ to act as a transparency control factor, and $\text{Aug}(\cdot)$ applies motion blur added using PR software. 

\begin{table}[htbp]
    \centering
    \renewcommand{\arraystretch}{1.2}
    \setlength{\tabcolsep}{5pt}
    \resizebox{\linewidth}{!}{%
        \begin{tabular}{c|c|c|c|c|c|c|c|c}
            \Xhline{2pt}
            \multirow{2}{*}{Dataset} 
            & \multirow{2}{*}{Type} 
            & \multirow{2}{*}{Resolution} 
            & \multirow{2}{*}{\shortstack{Size\\(frames)}} 
            & \multirow{2}{*}{\shortstack{Smoke\\Types}} 
            & \multirow{2}{*}{\shortstack{Source\\Detection}} 
            & \multirow{2}{*}{Turbulence} 
            & \multirow{2}{*}{\shortstack{Cavity\\Pressure}} 
            & \multirow{2}{*}{\shortstack{Scenario\\Sources}} \\
            & & & & & & & & \\
            \Xhline{1pt}
            MARS-GAN & Image & 256$\times$256 & 18000 & 1 & \ding{55} & \ding{55} & \ding{55} & 1 \\
            PFAN & Image & 480$\times$480 & 660 & 1 & \ding{55} & \ding{55} & \ding{55} & 1 \\
            PSv2rs & Image & 256$\times$256 & 54,420 & 1 & \ding{55} & \ding{55} & \ding{55} & 1 \\
            \cdashline{1-9}
            \textbf{Ours}   & \textbf{Video} & \textbf{720$\times$1080} & \textbf{12,000} & \textbf{3} & \cmark & \cmark & \cmark & \textbf{3} \\
            \Xhline{2pt}
        \end{tabular}%
    }
    \caption{Comparison with recent synthetic desmoking datasets \cite{hong2023mars,zhang2023progressive, liu2024smoke}.}
    \label{tab:dataset_comparison}
\end{table}

\begin{figure*}[t]
  \centering

  \includegraphics[width=\textwidth]{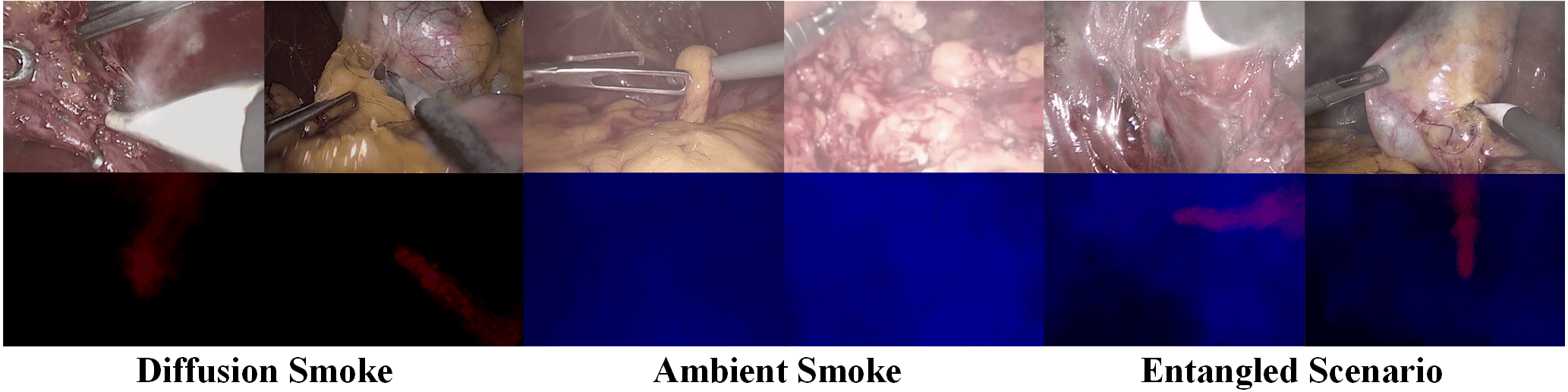}
  \caption{Examples of diverse scenarios and smoke types in the STSVD dataset.}
  \label{d_vis}

  \vspace{2em}

  \includegraphics[width=\textwidth]{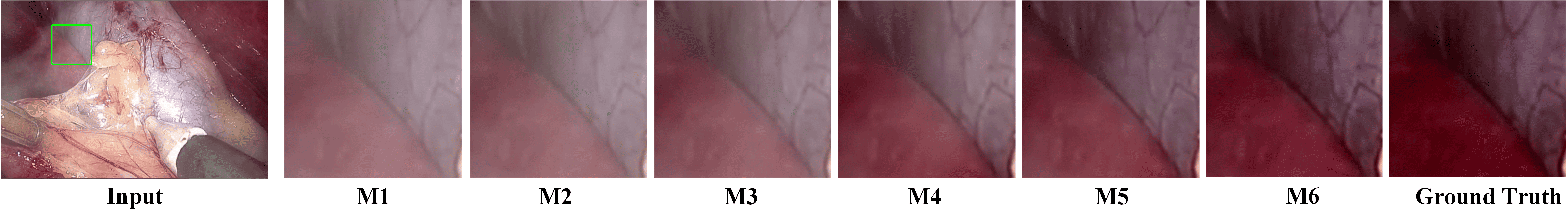}
  \caption{The qualitative comparisons of ablation study in STSVD dataset.}
  \label{fig_a_syn}

  \vspace{2em}

  \includegraphics[width=\textwidth]{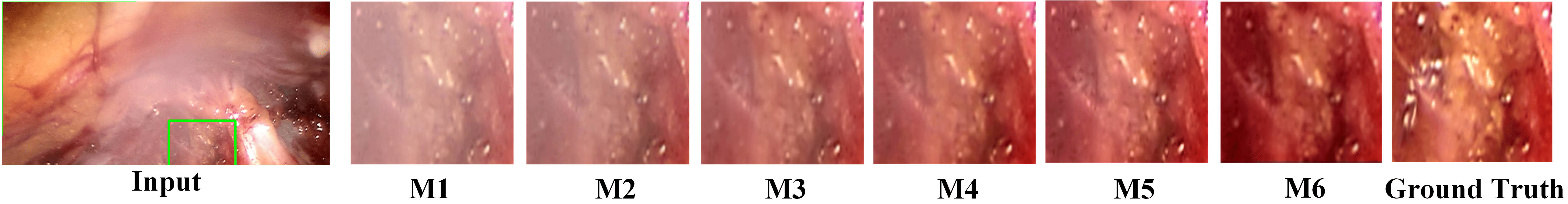}
  \caption{The qualitative comparisons of ablation study in Vivo dataset.}
  \label{fig_a_vivo}

  \vspace{2em}

  \includegraphics[width=\textwidth]{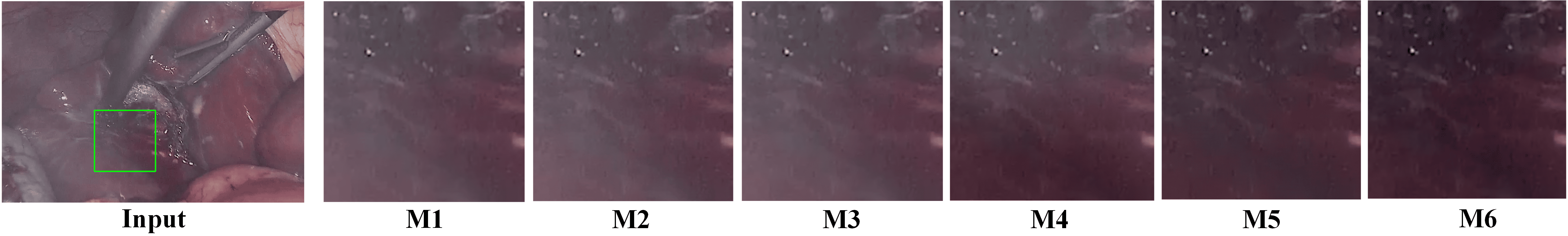}
  \caption{The qualitative comparisons of ablation study in SVSTD-R dataset.}
  \label{fig_a_real}

\end{figure*}

\subsubsection{Dataset Comparison and Visualization}

To evaluate the comprehensiveness of datasets, we compare our proposed synthetic dataset against several recent alternatives, as summarized in Tab\ref{tab:dataset_comparison}. Unlike previous datasets which only provide image-level samples with a single smoke type, our dataset introduces temporally consistent video sequences with three distinct smoke types. Meanwhile, we adopt a more refined rendering pipeline, incorporating tool tip detection, turbulence dynamics, and cavity pressure conditions, which enables the generated smoke masks to better resemble real surgical smoke. As shown in Fig.\ref{d_vis}, we present sample visualizations of the STSVD dataset across various scenarios, including Diffusion Smoke, Ambient Smoke, and entangled scenarios. For better visual distinction, red and blue pseudo-colors are used to represent Diffusion smoke and Ambient Smoke, respectively.

\section{B.More Qualitative Ablation Results}

As shown in Fig.\ref{fig_a_syn}, Fig.\ref{fig_a_vivo} and Fig.\ref{fig_a_real}, we present additional qualitative ablation results on STSVD, Vivo, and SVSTD-R datasets, respectively.

\section{C.More Qualitative Comparison Results}

\begin{figure*}[t]
  \centering

  \includegraphics[width=0.9\textwidth]{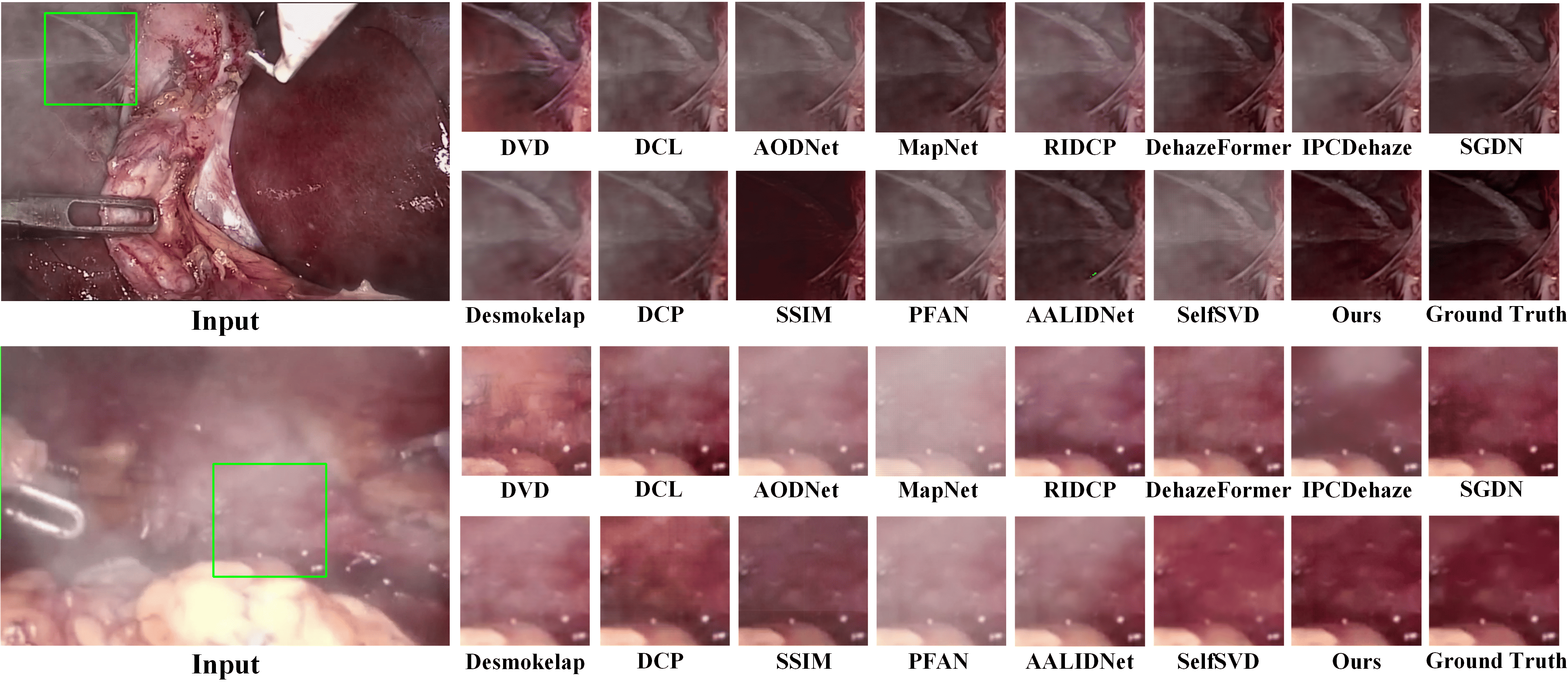}
  \caption{The qualitative comparison between different methods in STSVD dataset.}
  \label{fig_com_syn}

  \vspace{1em}

  \includegraphics[width=0.9\textwidth]{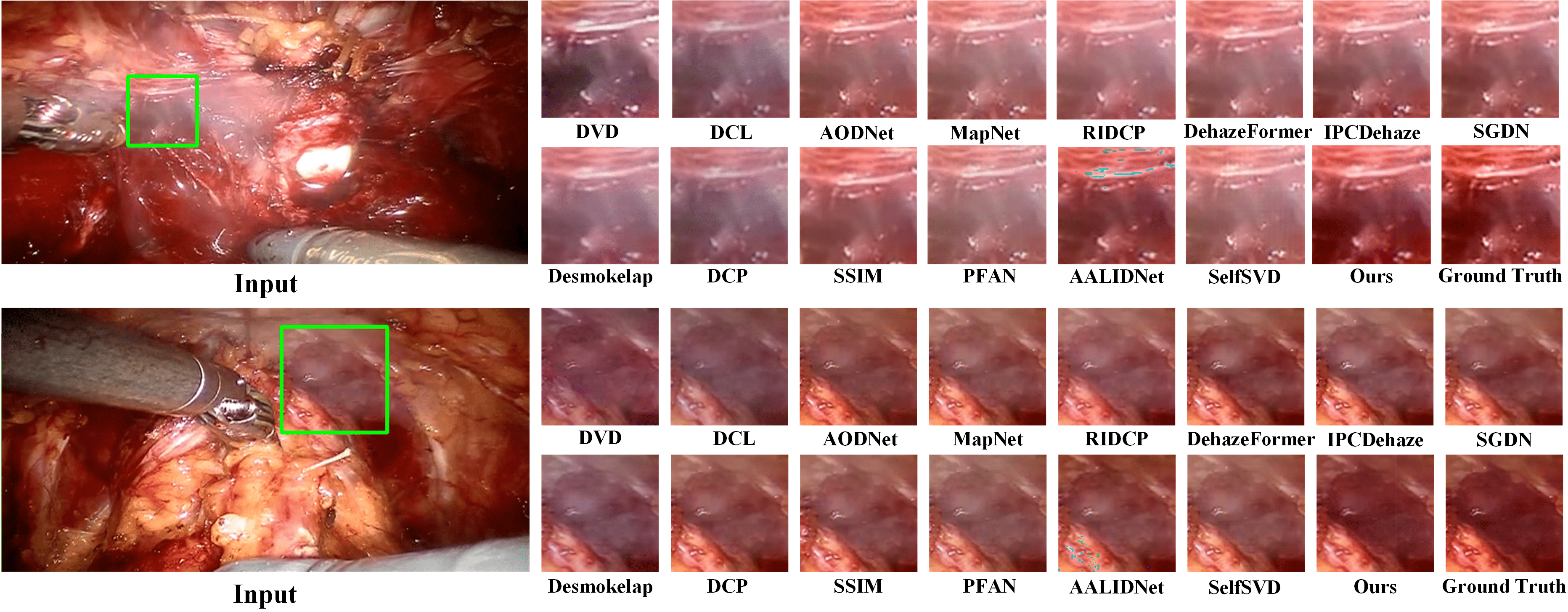}
  \caption{The qualitative comparison between different methods in Vivo dataset.}
  \label{fig_com_vivo}

  \vspace{1em}

  \includegraphics[width=0.9\textwidth]{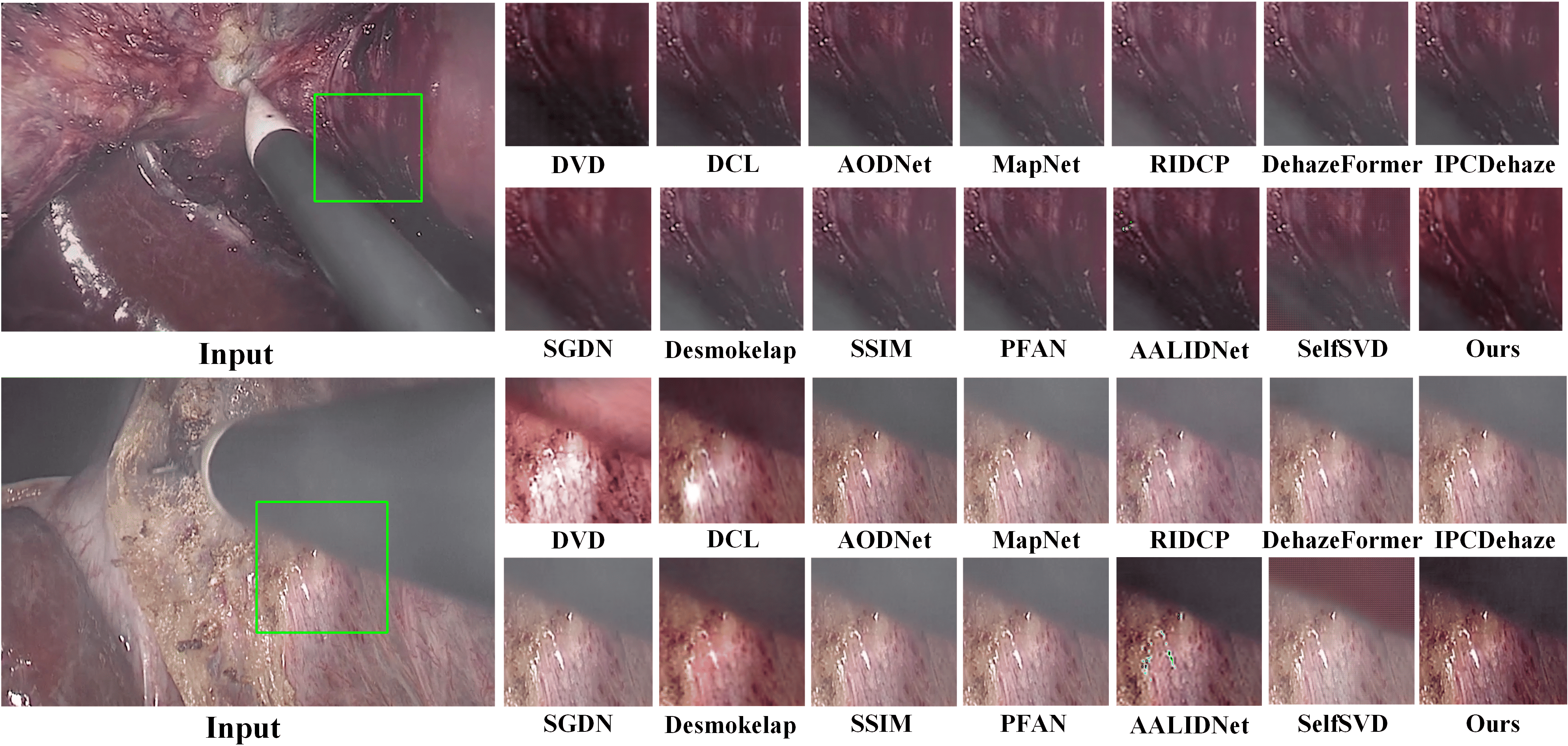}
  \caption{The qualitative comparison between different methods in STSVD-R dataset.}
  \label{fig_com_real}

  \vspace{1em}

\end{figure*}

As shown in Fig.\ref{fig_com_syn}, Fig.\ref{fig_com_vivo} and Fig.\ref{fig_com_real}, we present additional qualitative comparison results on STSVD, Vivo, and SVSTD-R datasets, respectively.

\end{document}